\documentclass{article}
\usepackage{mlspconf}
\usepackage{cite}
\usepackage{amsmath,amssymb,amsfonts}
\usepackage{algorithmic}
\usepackage{graphicx}
\usepackage{textcomp}
\usepackage{xcolor}
\usepackage{amsmath}

\usepackage{algorithmic}
\usepackage{graphicx}
\usepackage{textcomp}
\usepackage{xcolor}
\usepackage{float}
\usepackage{wrapfig}
\usepackage{array}
\usepackage{multirow}
\usepackage{hyperref}
\usepackage{comment}
\usepackage{enumitem}  
\hypersetup{
    colorlinks=true,
    linkcolor=blue,
    filecolor=magenta,      
    urlcolor=blue,
}

\newcolumntype{L}[1]{>{\raggedright\let\newline\\\arraybackslash\hspace{0pt}}m{#1}}
\newcolumntype{C}[1]{>{\centering\let\newline\\\arraybackslash\hspace{0pt}}m{#1}}
\newcolumntype{R}[1]{>{\centering\let\newline\\\arraybackslash\hspace{0pt}}m{#1}}

\usepackage{lipsum}

\let\OLDthebibliography\thebibliography
\renewcommand\thebibliography[1]{
  \OLDthebibliography{#1}
  \setlength{\parskip}{0pt}
  \setlength{\itemsep}{0pt plus 0.3ex}
}

\title{Identifying Stochasticity in Time-Series with Autoencoder-Based Content-aware 2D Representation: Application to Black Hole Data}
\name{Chakka Sai Pradeep, Neelam Sinha\thanks{We are thankful to Prof. Banibrata Mukhopadhyay, Department of Physics, Indian Institute of Science Bangalore, for sharing the data and helping in formulating the problem statement.}}
\address	{International Institute of Information Technology\\
	Bangalore 560100, India\\
	{\tt\small \{saipradeep.chakka, neelam.sinha\}@iiitb.ac.in}}

\begin{document}

\maketitle

\begin{abstract}
In this work, we report an autoencoder-based 2D representation to classify a time-series as stochastic or non-stochastic, to understand the underlying physical process. Content-aware conversion of 1D time-series to 2D representation, that simultaneously utilizes time- and frequency-domain characteristics, is proposed. An  autoencoder is trained with a loss function to learn latent space (using both time- and frequency domains) representation, that is designed to be, time-invariant. Every element of the  time-series is represented as a tuple with two components, one each, from latent space representation in time- and frequency-domains, forming a binary image. In this binary image, those tuples that represent the points in the time-series, together form the ``Latent Space Signature" (LSS) of the input time-series. The obtained binary LSS images are fed to a classification network. The EfficientNetv2-S classifier is trained using 421 synthetic time-series, with fair representation from both categories. The proposed methodology is evaluated on publicly available astronomical data which are 12 distinct temporal classes of time-series pertaining to the black hole GRS 1915 + 105, obtained from RXTE satellite. Results obtained using the proposed methodology are compared with existing techniques. Concurrence in labels obtained across the classes, illustrates the efficacy of the proposed 2D representation using the latent space co-ordinates. The proposed methodology also outputs the confidence in the classification label.
\end{abstract}
\begin{keywords}
Autoencoder, black hole time-series, stochasticity, time-invariance, latent space representation, classification
\end{keywords}
\vspace{-1.5em}
\section{Introduction}
\label{intro}







In order to carry out any study, across multiple domains, such as weather, finance, agriculture, astronomy, researchers analyse measurements and create models aimed at explaining the underlying phenomena. Several of these measurements are captured as time-series that need to be carefully examined. The first step towards examining the time-series is to determine if the time-series is stochastic, which would mean that it is noise; or if the time-series is non-stochastic, in which case physical attributes explaining them would be essential. As is well-studied, a stochastic time-series is one in which the  sequence of independent realizations of a random variable are not related.  Standard examples of well-studied stochastic time-series are white noise, pink noise, etc. On the other hand, standard examples of non-stochastic time-series such as logistic map (at growth rate = 4), Lorenz system  result in time-series that exhibit a well-defined structure.

An important problem in astronomy is the study of black hole sources. The challenge is that to identify a black hole, one needs to study its environment often a disc-like structure formed due to the in-falling matter called accretion disc. The time-series of black hole source, measured at different times, could exhibit drastically varying behavior, at times being  stochastic and other times being non-stochastic. Based on their characteristics, they are modeled differently and sub-categorised for further analysis.  Here, we study the black hole source \textit{GRS 1915+105}, which has 12 different temporal classes: $\alpha$, $\beta$, $\gamma$, $\delta$, $\lambda$, $\kappa$, $\mu$, $\nu$, $\rho$, $\phi$, $\chi$ and $\theta$. The different temporal classes represent the different time-series measured from the same source that exhibits different characteristics at different times. The differences across these classes lie in level of X-ray emission, particularly in different energy bands. Also they exhibit different combinations of QPO frequencies and PSD structure. The study in \cite{Belloni} explains the differences across the temporal classes and their implications.

\begin{figure*}[h]
  \centering
  \includegraphics[width=\linewidth]{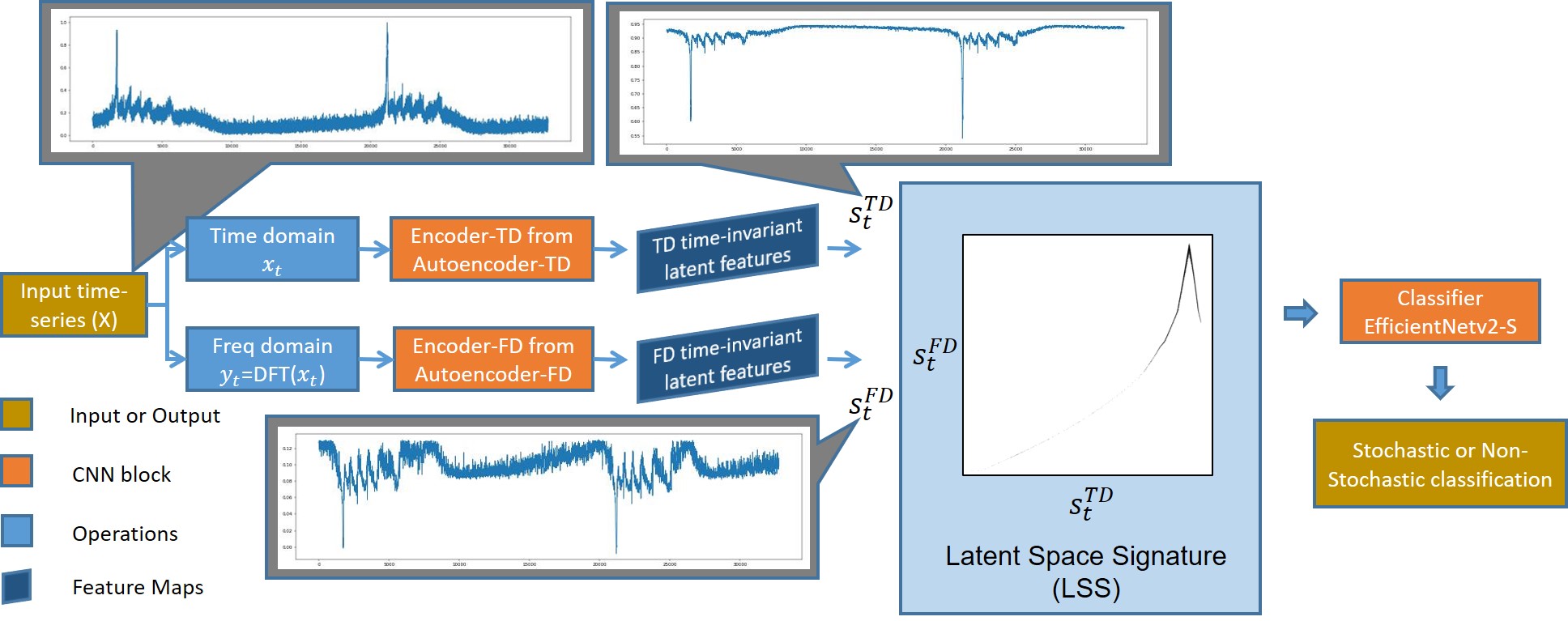}
  \caption{System Overview Illustrated on a representative real time-series ($\alpha$- temporal class) }
  \label{overview}
\end{figure*}

The most popular method to study a time-series, is by determining the Correlation Dimension (CD) for varying values of Embedding Dimension (ED), by using the Correlation Integral (CI) approach, as outlined in \cite{CIGRacia}. Several variants \cite{chris_1915} of this approach have emerged over the years. In order to declare a time-series as stochastic, the CI approach computes CD for increasing values of ED. If it turns out that CD does not saturate for increasing values of ED, then the time-series is declared as stochastic, else it is considered as non-stochastic. The value of CD helps in understanding the number of independent factors required to understand the phenomenon giving rise to the time-series. However, this approach is known to be computationally expensive, since it needs to be repeated for various values of ED, before anything conclusive can be declared. The other methods in literature are based on concepts such as Entropy and Mutual Information. Entropy-based approaches \cite{splrecent}, \cite{russian}, \cite{Boaretto2021} relied on ideas of phase-space reconstruction, recurrence plots, which inherently required domain knowledge. The work reported in \cite{Boaretto2021} parameterized a given time-series and utilized Permutation Entropy (PE), in order to enable Neural-Network based classification. PE was used to determine the measure of resemblance with known stochastic signals such as pink noise. The claim was that for non-stochastic signals the deviation of the parameter is relatively large as compared to that of the parameter of a stochastic signal. The authors reported results on synthetic signals such as various types of noise, chaotic systems as well as empirical data sets such as human gait data and heart rate variability data. This approach, however, assumes prior knowledge on the length of ordinal sequences.

An entirely different set of approaches utilized an alternate representation of the time-series, such as a graph. The work in \cite{lacasa2010} utilized horizontal visibility algorithm to classify a time-series. Yet another reported work \cite{Silva2022} mapped  time-series to graphs followed by computing various topological properties, which they called \textit{NetF}, capturing  measures such as centrality, distance, connectivity etc. Clustering was carried out on these measures, leading to the desired labelling of the time-series. The study reported in \cite{Brunton2016} determined the dynamics by exploring ideas of sparsity and machine learning in non-linear dynamical systems. Of late, deep learning for time-series classification, as summarized in the survey paper \cite{review2019paper}, is being explored by several research groups. Generative models such as Autoencoders are used, along with CNNs and LSTMs. Discriminative models based on architectures such as ResNet, Variations of LeNet, etc are also explored mainly for efficacy in feature engineering. 

Another set of approaches in time-series analysis relies on converting it to a 2D representation \cite{Wang2014EncodingTS}. The image obtained using the 2D representation is used for subsequent analysis, such as classification. Popular methods to obtain the conversion include  Markov Transition Fields and Grammian Angular Fields. These approaches utilize the temporal ordering but lack content-awareness and hence are not robust to noise. Besides, the data to be processed using these conversions increase exponentially with the length of the time-series, making them unscalable. 

In this work, we utilize autoencoders to obtain a novel representation of the 1D time-series, that is utilized to construct 2D binary images. The binary images are constructed by utilizing 2D representation of the time-series in latent space learnt by autoencoders. These autoencoders are trained to obtain time-invariant features by analyzing the input time-series in both time- and frequency-domains, separately. These binary images are henceforth referred as ``Latent Space Signatures" (LSS) of the time-series, which are utilized for classification. Besides the label, the confidence in the label is also measured.






\vspace{-1.5em}
\section{Proposed Method}
\label{proposed_method}
The aim of this work is to determine if the given 1D time-series is stochastic or non-stochastic. This helps in understanding the phenomenon that gives rise to the time-series.


Towards this, contributions in this paper are: 
(i) Usage of autoencoder \cite{deep_learning} for obtaining novel latent space based content-aware 2D representation of a 1D time-series, here called ``Latent Space Signature" (LSS). (ii)  Utility of LSS to determine if a time-series is stochastic or non-stochastic. (iii) Novel DNN-based methodology for classifying a 1D time-series as stochastic or Non-stochastic. (iv) Confidence in label can be interpreted as a reflection of SNR of the 1D time-series.
 
In this work, we propose the usage of autoencoder based conversion of 1D time-series to 2D representation in latent-space obtained by training for time-invariant features, in both time- and frequency domains, separately. Fig. \ref{overview} describes the proposed system overview. Dataset and code repository is provided at: \url{https://github.com/csai-arc/blackhole_1D_2D_label}.

\begin{figure}[h]
  \centering
  \includegraphics[width=\linewidth]{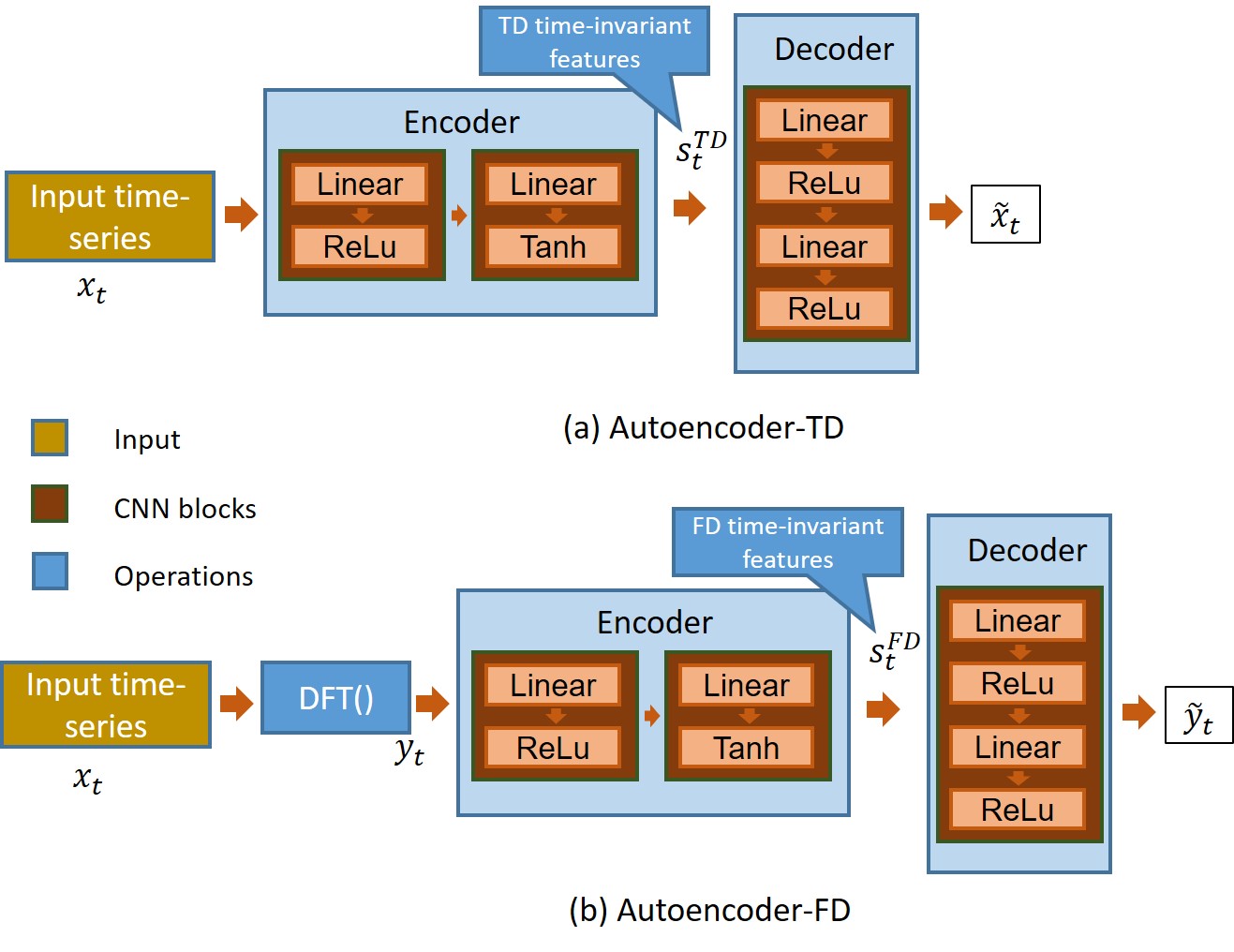}
  \caption{Autoencoder Training Methodology}
  \label{training}
\end{figure}

\vspace{-1.5em}
\subsection{Autoencoder Training Methodology:}
\begin{figure}[h]
  \centering
  \includegraphics[width=\linewidth]{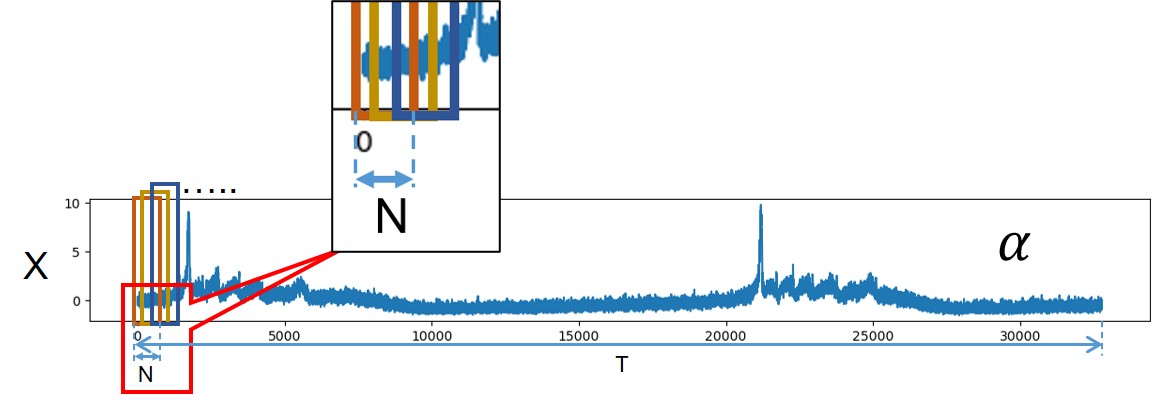}
  \caption{Window size $N$ and time-series length $T$ on black hole source $\alpha$ time-series (X)}
  \label{X_ts_division}
\end{figure}

Overview of the proposed autoencoder training methodology is shown in Fig. \ref{training}. Let $X$ be a time-series of length $T$ as in Fig. \ref{X_ts_division}. We divide $X$ into windows of size $N $ $(N=10)$ and obtain $x_{t}=\left [ X[t-N+1], ...X[t] \right ]^{T}$ where  
$x_{t}\in \mathbb{R}^{N}$.

Also, we compute discrete Fourier transform (DFT) on each window $x_{t}$ to obtain spectral information. Further, we perform the following operations: (a) Crop the transformed window to a predefined length of $M$ (here we use $M=N$) and (b) compute the modulus of the transformed window. We represent DFT operation along with the operations (a) and (b) as $\mathbb{F}=\mathbb{R}^{N}\rightarrow \mathbb{R}^{M}$. We represent this operation as $y_{t}= \mathbb{F}(x_{t}) $ where $y_{t}\in \mathbb{R}^{M}$. Hence this represents the frequency-domain content in the given time-series.

We propose usage of autoencoders (AEs), extending the works reported in \cite{tire}, \cite{tire_back}, for time-invariant feature extraction in both time-domain (TD) and frequency-domain (FD), $x_{t}$ and $ y_{t}$ respectively. AEs are trained over a loss function which promotes time-invariance over successive  windows. As illustrated in Fig. \ref{training} (a), TD encoder maps $x_{t}\in \mathbb{R}^{N}$ to its encoded form. 
Encoder is represented as:
\begin{align}
s_{t}^{TD}=\sigma (W_{TD}^{1}*x_{t}+b_{TD}^{1}) & & where \; 
s_{t}^{TD}\in \mathbb{R}^{s} 
\label{st}
\end{align}

where $W_{TD}$, $b_{TD}$ and $\sigma$ respectively represent weight, bias and non-linear activation function \cite{topology}. $s_{t}^{TD}$ represent time-invariant latent space features that are learnt using the time domain representation of the given time-series. Decoder is represented as:
\begin{align}
\tilde{x}_{t}={\sigma}' ({W}'_{TD}*s_{t}^{TD}+{b}'_{TD}) & & where \; 
\tilde{x}_{t}\in \mathbb{R}^{N}
\label{xtn}
\end{align}

Features learned by the autoencoder in the latent space, can be called time-invariant, if the change in them is minimal, along the entire span of the time-series. We use the following loss function to ensure that the latent-space features $s_{t}$ are time-invariant; also the reconstruction obtained is faithful to the original. The Loss function is given by,
\begin{align}
\mathbb{L}^{TD}=\sum_{t\in T_{j}}\left ( \left \| x_{t}-\tilde{x}_{t} \right \| + \frac{1}{K}\sum_{k=0}^{K-1}\left \|s_{t-k}^{TD}-s_{t-k-1}^{TD}\right \|\right )
\label{loss}
\end{align}
The first term, $\left \| x_{t}-\tilde{x}_{t} \right \|$, establishes faithful reconstruction of the original, whereas the second term, $\left \|s_{t-k}^{TD}-s_{t-k-1}^{TD}\right \|$, helps in obtaining time invariance in latent space features. We randomly partition for all time stamps $t$ over $J$ smaller mini batches $T_{j}$. $K$ indicates the number of consecutive features that are jointly considered for the computation of gradient $(K=3)$. This creates smoothing effect in the stochastic gradient descent optimization.

Similarly, as illustrated in Fig. \ref{training} (b), FD encoder maps $y_{t}\in \mathbb{R}^{M}$ to its encoded form $h_{t}^{FD}\in \mathbb{R}^{h}$. Encoder is represented as:
\begin{align}
s_{t}^{FD}=\sigma (W_{FD}^{1}*y_{t}+b_{FD}^{1}) & & where \; 
s_{t}^{FD}\in \mathbb{R}^{s} 
\label{stfd}
\end{align}
where $W_{FD}$, $b_{FD}$ and $\sigma$ respectively represent weight, bias and non linear activation function. $s_{t}^{FD}$ represent time-invariant latent space features learnt using the frequency domain representation of the input time-series. Decoder is represented as:
\begin{align}
\tilde{y}_{t}={\sigma}' ({W}'_{FD}*s_{t}^{FD}+{b}'_{FD}) & & where \; 
\tilde{y}_{t}\in \mathbb{R}^{M}
\label{xtnfd}
\end{align}

We use similar loss function as in equation (\ref{loss}) to make $s_{t}^{FD}$ features time-invariant and for faithful reconstruction, given by,
\begin{align}
\mathbb{L}^{FD}=\sum_{t\in T_{j}} \left ( \left \| y_{t}-\tilde{y}_{t} \right \| + \frac{1}{K}\sum_{k=0}^{K-1}\left \|s_{t-k}^{FD}-s_{t-k-1}^{FD}\right \|\right )
\label{lossfd}
\end{align}


Once the model is trained, $s_{t}^{TD}$ and $s_{t}^{FD}$ time-invariant features which are obtained from Encoder-TD and Encoder-FD respectively, are further processed. For each point in the given time-series, we obtain an alternate representation, which is a tuple of numbers from the time-domain latent space and the frequency-domain latent space, as explained below. 

\begin{figure}[h]
  \centering
  \includegraphics[width=\linewidth]{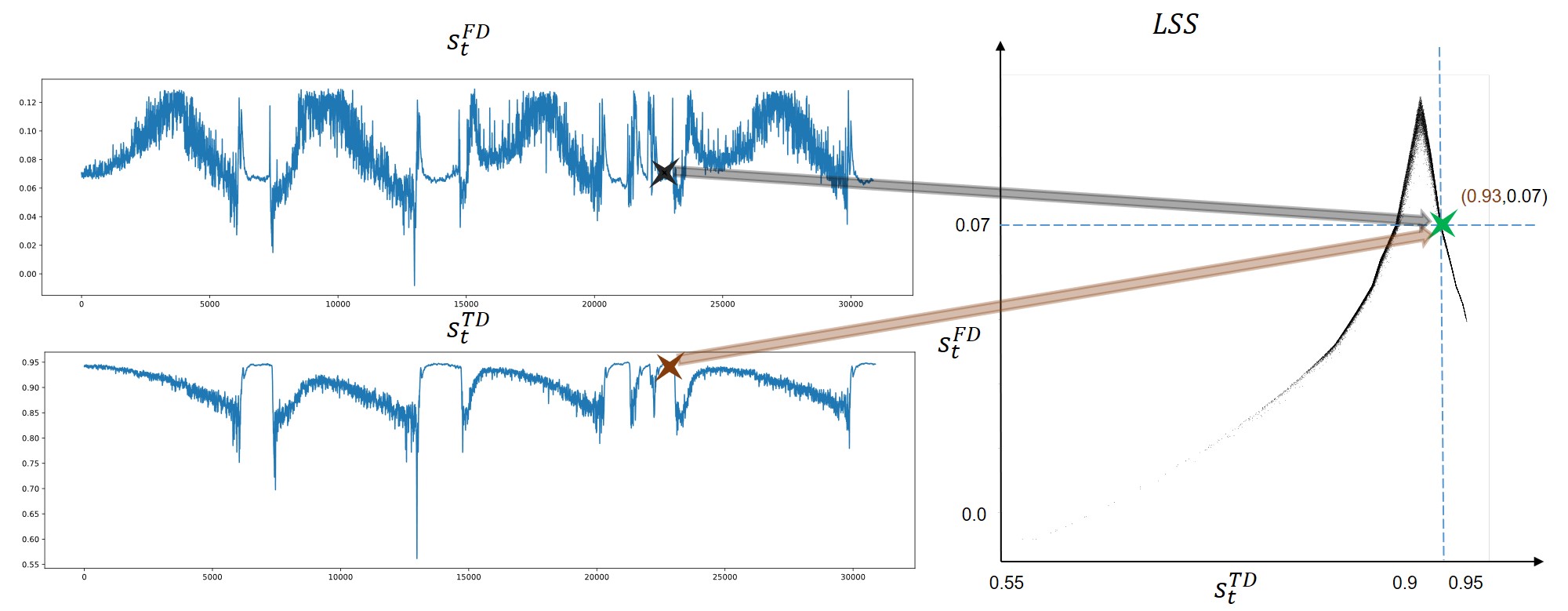}
  \caption{Forming the LSS image using the Latent-space representation obtained using Time- and Frequency-domain analyses is shown on a sample time-series (Theta)}
  \label{LSS_form}
\end{figure}

\vspace{-1.5em}
\subsection{Obtaining the Latent Space Signature (LSS):}
	In order to obtain the LSS image, the time-invariant latent space features, obtained using both time- and frequency-domain analyses carried out by the Autoencoders are utilized. For time-series $X$ of length $T$, window size of $N$ where $N<<T$ is used. Here $T$ is of the order of 10's of 1000's and $N$ utilized is 10.

$X = \{ x_{1} \hdots x_{t} \hdots  x_{T} \}$ is the input time-series. We propose a mapping that takes the scalar $x_{t}$ to a two dimensional vector representation given by: 
$x_{t}\rightarrow \begin{bmatrix}
s_{t}^{TD}\\ 
s_{t}^{FD}
\end{bmatrix}$
\begin{align}
LSS(X) = scatter(\begin{bmatrix}
s_{t}^{TD}\\ 
s_{t}^{FD}
\end{bmatrix}) \;
\label{LSS}
\end{align}
where $t \in \{1, \hdots, (T-N)\}$. This LSS image formation is illustrated in Fig. \ref{LSS_form}. Hence for every index in the original time-series, we obtain a tuple, one representing the time-domain content, and another representing the frequency-domain content. The latent space representation in time-domain, is used as the Horizontal axis, while the latent space representation in frequency-domain, is used as the Vertical axis. The new axes set is used to create a binary image. The binary image, with the afore-mentioned axes picked up from the latent space representation, is marked with those tuples that correspond to the elements in the time-series. This creates the LSS image of the given time-series. An illustration of the process is shown in Fig. \ref{LSS_form} for a representative time-series. As illustrated for a sample point in the LSS binary image whose co-ordinates are (0.93, 0.07), the value of 0.93 comes from the latent space representation in time domain (color coded brown), while the value of 0.07 comes from the latent space representation in frequency domain (color coded black), for a fixed time-stamp in the original time-series. These LSS images are seen to contain distinct patterns for stochastic and non-stochastic time-series, enabling the classification.


\vspace{-1.5em}
\subsection{Classifier Training Methodology:} For classification of LSS images, EfficientNetv2-S \cite{efficientnetv2} network architecture is trained using cross entropy loss. EfficientNetv2-S \cite{efficientnetv2} uses network architecture search (NAS) to design an optimal model. This architecture uses compound scaling which allows the network to handle images of various sizes and making it scale-invariant. Additionally, this architecture is computationally efficient and has small memory footprint.



We utilize cross-entropy loss for LSS image classification as below ($L_{CE}$):
\begin{align}
  L_{CE}  & =  -\gamma_{z}\cdot log\left ( \frac{e^{\hat{z}{{c}'}}}{\sum_{c}e^{\hat{z}^{c}}} \right )  \; 
\label{ppe_l_ce}
\end{align}
Here, $\hat{z}^{{c}'}$ denotes the CNN prediction score of the positive class. $\sum_{c}$ denotes sum of predictions over all classes. We set $\gamma_{z}=1$, for equal class-wise weights. On completion of training, during inference, softmax operation is performed on the final fully connected layer to obtain the prediction confidence scores $C_S$ (confidence in stochastic label) and $C_{NS}$ (confidence in non-stochastic label) for a given input LSS image. Based on $C_S$ and $C_{NS}$ scores, the one having the higher score is used to determine the LSS label.

\vspace{-0.4cm}
\section{Dataset}
In the training phase, the proposed algorithm utilizes synthetically generated time-series data (comprising of Lorenz, logistic maps, white noise and colored noise). The trained model is then evaluated on publicly available RXTE satellite data \cite{rxte} for the black hole source \textit{GRS 1915 + 105}. 12 temporal classes of time-series, re-sampled with sampling interval of 0.1 second, are utilized. The length of \textit{GRS 1915 + 105} time-series, considered across 12 temporal classes, varies from a minimum of 16000 to a maximum of 34000. The length of each synthetically generated time-series data (Lorenz, logistic maps, white noise and colored noise) is 30000. \textit{GRS 1915 + 105} dataset was also used earlier \cite{adigoke}, where the authors reported CI-based results.

\vspace{-1em}
\section{Results and Discussion}
\label{results}
We train our autoencoders for time-series reconstruction as illustrated in Fig. \ref{training} on synthetic data such as Lorenz, logistic maps, white noise and colored noise. After the Autoencoder training, time-invariant latent space representation is obtained using both time- and frequency domains. These latent space representations are utilized to generate the LSS binary images for each of the time-series. These LSS images are fed to the EfficientNetv2-S based classifier to obtain the classification labels. It must be noted that the classifier also outputs the confidence measure for each of the labels.
\vspace{-0.3cm}
\begin{table}[h]
	\centering
	\small
		\caption{Comparison of LSS Label On Black Hole Source \textit{GRS 1915 + 105} Data with CI Label}
		\centering
		\label{sm_real}
		\begin{tabular}{| C{1.6cm}| C{1cm} | C{1cm} | C{0.8cm} | C{0.8cm} | }
			\hline
			\textbf{Time-series} &\textbf{$C_S$} &\textbf{$C_{NS}$} &LSS Label &CI Label\\ \hline
				$\chi$			&0.9898	&0.0102 &S &S \\ \hline
				$\gamma$		&1.00	&0.00 &S &S \\ \hline
				$\phi$			&1.00	&0.00 &S &S \\ \hline
				$\delta$&0.807	&0.193 &S&S \\ \hline
				$\mu$			&0.0041	&0.9959 &NS &NS \\ \hline
				$\nu$			&0.0779	&0.9221 &NS &NS \\ \hline
				$\alpha$		&0.0055 &0.9945 &NS &NS \\ \hline
				$\theta$		&0.1206	&0.8794 &NS &NS \\ \hline
				$\rho$			&0.1593	&0.8407 &NS &NS \\ \hline
				$\beta$			&0.0083	&0.9917 &NS &NS \\ \hline
				$\kappa$		&0.0102	&0.9898 &NS &NS \\ \hline
				$\lambda$		&0.0043	&0.9957
				 &NS &NS \\ \hline
		\end{tabular}                                              
\end{table}

Table \ref{sm_real} shows our proposed methodology performance on black hole source \textit{GRS 1915 + 105} data. We compare our results with those obtained using CI method \cite{adigoke}, and obtain concurrence on all the classes. Additionally, we provide the value of the confidence scores $C_{S}$ and $C_{NS}$, leading to richer understanding. We also compute Class Activation Maps to illustrate the explainability in classification.

\vspace{-1em}
\subsection{Latent Space Signature (LSS) Images:}
\begin{figure}[h]
	\centering
	\includegraphics[width=0.95\linewidth]{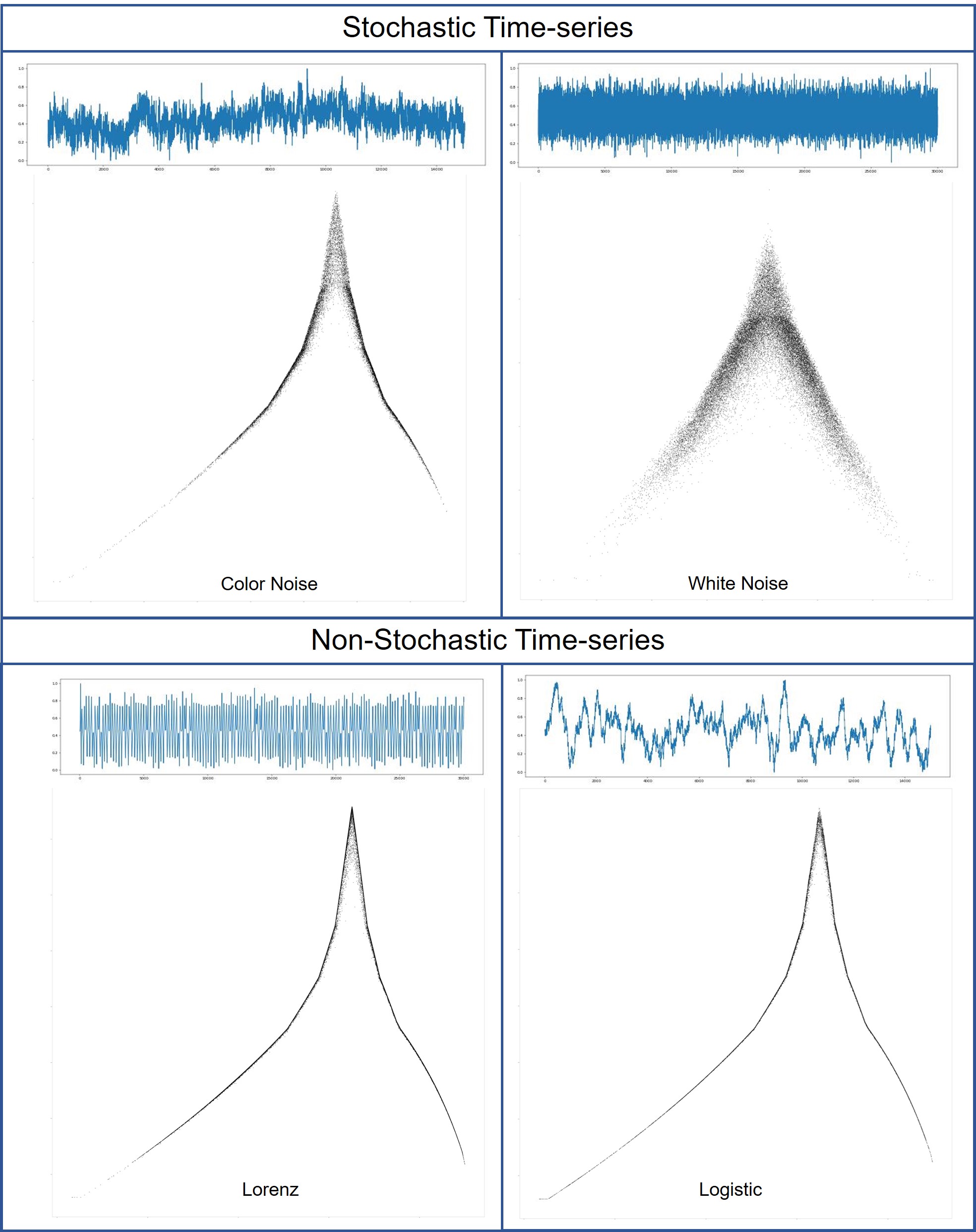}
	\caption{The distinct characteristics in LSS images corresponding to stochastic and non-stochastic labels are compared}
	\label{LSS_S_NS}
\end{figure}

We train our EfficientNetv2-S classifier for time-series classification on synthetic LSS data. We generated 210 non-stochastic (Lorenz, Logistic map) LSS synthetic images and 211 stochastic (White noise, Color noise) LSS synthetic images. Representative time-series and their respective LSS images are shown in Fig. \ref{LSS_S_NS}. The top panel - left shows a realization of color noise and its LSS image; the top panel - right shows a realization of white noise and its LSS image. Similarly, the bottom panel - left shows a realization of time-series obtained from Lorenz system and its LSS image; the bottom panel - right shows a realization of logistic map (growth rate = 4) and its LSS image. 

Visual examination shows that LSS images for non-stochastic label are sharp, while those for stochastic are spread out. This points to the hypothesis that for non-stochastic time-series multiple elements are mapped to the same tuple repeatedly in the latent space representation. On the other hand, for stochastic time-series, elements are mapped to more distinctive sets of tuples in the latent space representation. Yet another perspective is that, if the LSS images are interpreted as matrices, the ones corresponding to non-stochastic time-series will be sparser compared to the ones corresponding to stochastic time-series.

\begin{figure}[h]
	\centering
	\includegraphics[width=0.95\linewidth]{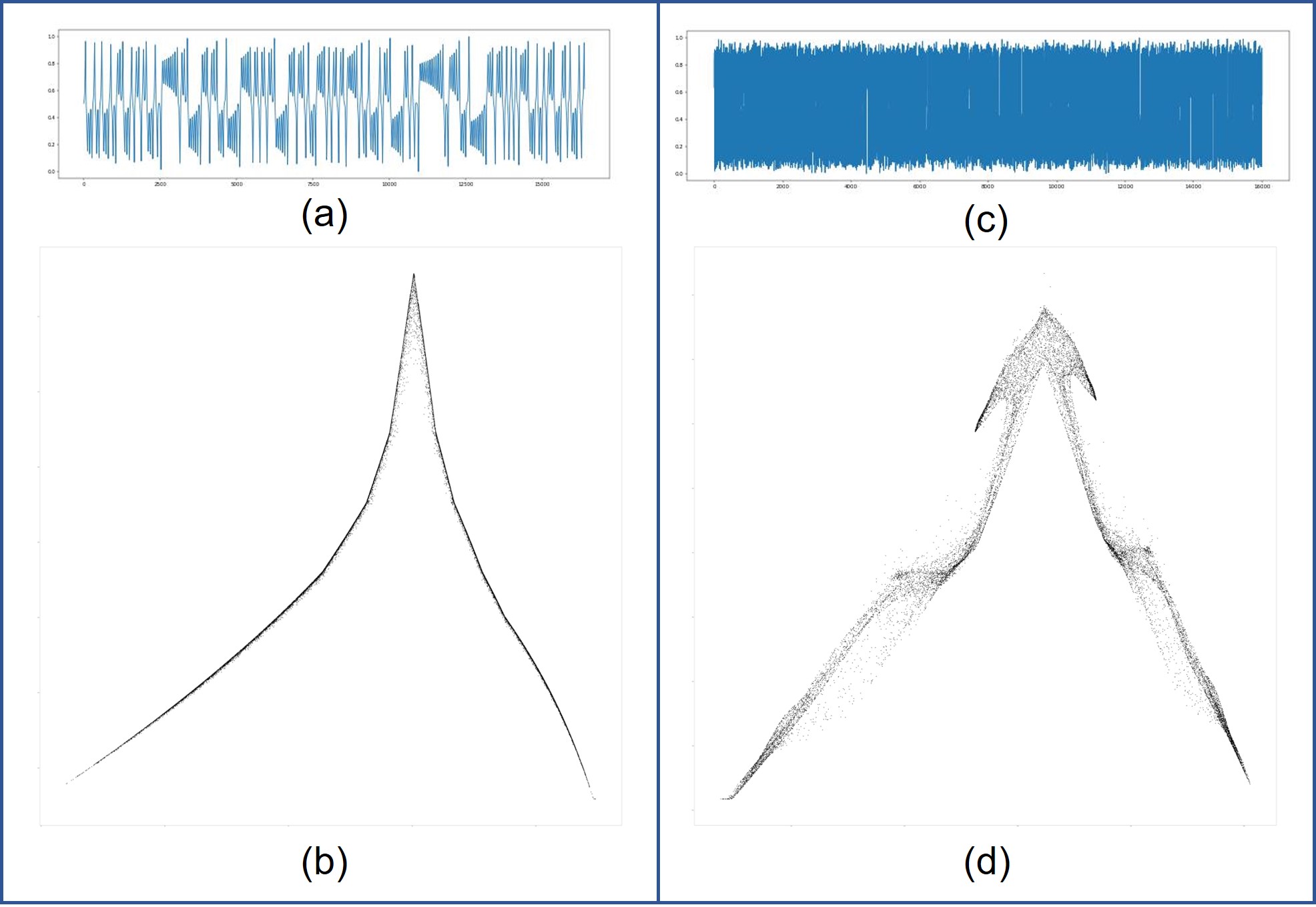}
	\caption{(a) Lorenz time-series, (b) LSS of Lorenz time-series, (c) Noise-corrupted Lorenz time-series, (d) LSS of noise-corrupted Lorenz time-series}
	\label{lorentz_S_NS}
\end{figure}

Fig. \ref{lorentz_S_NS} shows a specific example that illustrates the variations in LSS images. Fig. \ref{lorentz_S_NS} (a) and Fig. \ref{lorentz_S_NS} (b) represent Lorenz time-series and its corresponding LSS image respectively. Fig. \ref{lorentz_S_NS} (c) represents noisy lorenz, the noise-corrupted version of the time-series in Fig. \ref{lorentz_S_NS} (a). Fig. \ref{lorentz_S_NS} (d) represents LSS image of noisy lorenz shown in Fig. \ref{lorentz_S_NS} (c).

The comparison of LSS images shown in Fig.\ref{LSS_S_NS} and Fig. \ref{lorentz_S_NS} establishes the distinction in characteristics in the time-series corresponding to the two labels. 

\begin{figure}[h]
	\centering
	\includegraphics[width=0.9\linewidth]{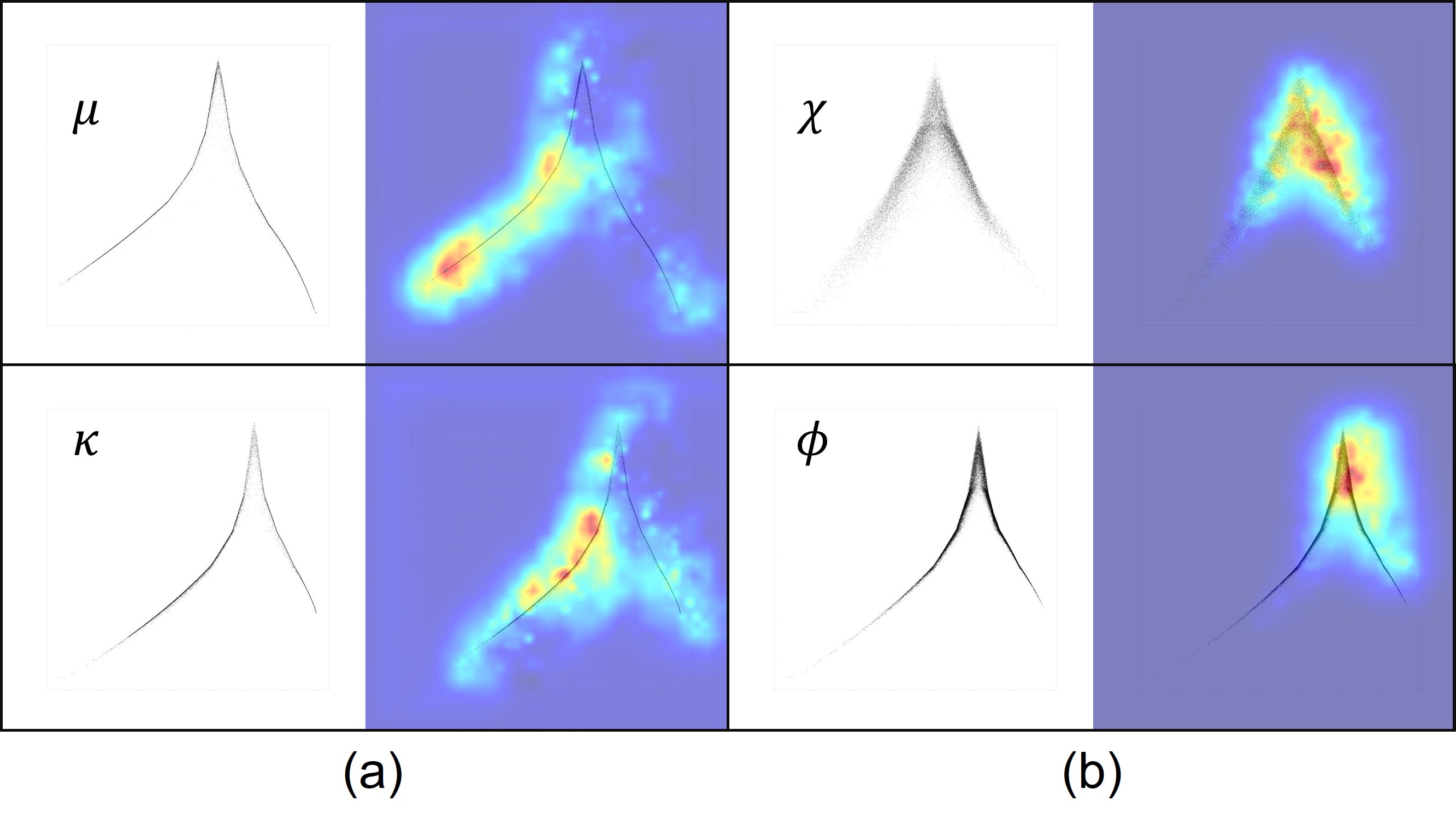}
	\caption{Class Activation Maps for real data - Left panel (a) corresponding to non-stochastic label, Right panel (b) corresponding to stochastic label.}
	\label{cams}
\end{figure}

\vspace{-1.0em}
\subsection{Class Activation Maps} 
The Class Activation Maps provided in Fig. \ref{cams} shows the regions in the LSS images that the classifier utilizes to decide on the label of the time-series. From the figure, it is clear for the representative two temporal classes of non-stochastic label, shown on left, called $\mu$ and $\kappa$, the sharp rising edge of the LSS is most significant, indicated by the bright red shades in the activation map. On the other hand, for the representative two temporal classes of stochastic label, shown on right, $\chi$ and $\phi$, the region that captures the spreading of values is most significant, shown in bright red shades in the corresponding activation maps. Similar trends are observed in all of the 12 temporal classes.

\vspace{-0.5cm}
\subsection{Comparison with CI methodology}
CI analyses the time-series by determining the CD to understand the underlying physical process. CI works by clustering of points with an appropriate choice of parameters that is chosen from the domain knowledge. CI method typically requires several iterations and possible computation of additional parameters such as Lyapunov exponent in order establish a time-series as stochastic or non-stochastic. However, the proposed methodology can identify the label in one single run, and also provide the confidence in the classification label. The proposed LSS images, along with CI approach can be used for exhaustive understanding of the time-series.

\vspace{-0.3cm}
\section{Conclusion}
\label{conclusion}
We propose a deep learning frame-work to convert a 1D time-series as a 2D representation, for the purpose of classification into stochastic or non-stochastic. Towards this, we utilize autoencoders that analyse the given time-series in both time- and frequency-domains. The latent space representations are utilized to create the binary LSS images. These LSS images are fed to a classifier to obtain the classification label corresponding to the time-series. The efficacy of the proposed methodology is illustrated on 12 temporal time-series classes of black hole \textit{GRS 1915+105} obtained from RXTE satellite data. We compare inferences of the CI-based approach with those obtained using the proposed methodology, and obtain concurrence on all the temporal classes.  
\vspace{-0.3cm}
\bibliographystyle{IEEEbib}
\bibliography{refs}

\begin{thebibliography}{10}

\bibitem{Belloni}
T~{Belloni}, M~{Klein-Wolt}, M~{Mendez}, M~{van der Klis}, and J~{van
  Paradijs},
\newblock ``{A model-independent analysis of the variability of GRS
  1915+105},''
\newblock {\em Astronomy \& Astrophysics}, vol. 355, pp. 271--290, Mar. 2000.

\bibitem{CIGRacia}
P~Grassberger and I~Procaccia,
\newblock {\em Measuring the Strangeness of Strange Attractors}, pp. 170--189,
\newblock Springer New York, New York, NY, 2004.

\bibitem{chris_1915}
C~Done, G~Wardziński, and M~Gierliński,
\newblock ``{GRS 1915+105: the brightest Galactic black hole},''
\newblock {\em Monthly Notices of the Royal Astronomical Society}, vol. 349,
  no. 2, pp. 393--403, 04 2004.

\bibitem{splrecent}
M~Rostaghi and H~Azami,
\newblock ``Dispersion entropy: A measure for time-series analysis,''
\newblock {\em IEEE Signal Processing Letters}, vol. 23, no. 5, pp. 610--614,
  2016.

\bibitem{russian}
L.~O Kirichenko, Y.~A Kobitskaya, and A.~Y Habacheva,
\newblock ``Comparative analysis of the complexity of chaotic and stochastic
  time series,''
\newblock {\em Radio Electronics, Computer Science, Control}, , no. 2, Nov.
  2014.

\bibitem{Boaretto2021}
B.~R.~R Boaretto, R.~C Budzinski, K.~L Rossi, T.~L Prado, S.~R Lopes, and
  C~Masoller,
\newblock ``Discriminating chaotic and stochastic time series using permutation
  entropy and artificial neural networks,''
\newblock {\em Sci Rep}, vol. 11, no. 1, pp. 15789, Aug. 2021.

\bibitem{lacasa2010}
L~Lacasa and R~Toral,
\newblock ``Description of stochastic and chaotic series using visibility
  graphs,''
\newblock {\em Phys. Rev. E}, vol. 82, pp. 036120, Sep 2010.

\bibitem{Silva2022}
V.~F Silva, M.~E Silva, P~Ribeiro, and F~Silva,
\newblock ``Novel features for time series analysis: a complex networks
  approach,''
\newblock {\em Data Mining and Knowledge Discovery}, vol. 36, no. 3, pp.
  1062--1101, May 2022.

\bibitem{Brunton2016}
S.~L Brunton, J.~L Proctor, and J.~N Kutz,
\newblock ``Discovering governing equations from data by sparse identification
  of nonlinear dynamical systems,''
\newblock {\em Proceedings of the National Academy of Sciences}, vol. 113, no.
  15, pp. 3932--3937, 2016.

\bibitem{review2019paper}
H~Ismail~Fawaz, G~Forestier, J~Weber, L~Idoumghar, and P.-A Muller,
\newblock ``Deep learning for time series classification: a review,''
\newblock {\em Data Mining and Knowledge Discovery}, vol. 33, no. 4, pp.
  917--963, Jul 2019.

\bibitem{Wang2014EncodingTS}
Z~Wang and T~Oates,
\newblock ``Encoding time series as images for visual inspection and
  classification using tiled convolutional neural networks,''
\newblock 2014.

\bibitem{deep_learning}
I~Goodfellow, Y~Bengio, and A~Courville,
\newblock {\em Deep Learning},
\newblock MIT Press, 2016,
\newblock \url{http://www.deeplearningbook.org}.

\bibitem{tire}
T.~D Ryck, M.~D Vos, and A~Bertrand,
\newblock ``Change point detection in time series data using autoencoders with
  a time-invariant representation,''
\newblock {\em CoRR}, vol. abs/2008.09524, 2020.

\bibitem{tire_back}
W~Lee, J~Ortiz, B~Ko, and R.~B Lee,
\newblock ``Time series segmentation through automatic feature learning,''
\newblock {\em CoRR}, vol. abs/1801.05394, 2018.

\bibitem{topology}
G~Naitzat, A~Zhitnikov, and L.-H Lim,
\newblock ``Topology of deep neural networks,''
\newblock {\em J. Mach. Learn. Res.}, vol. 21, no. 1, jun 2022.

\bibitem{efficientnetv2}
M~Tan and Q~Le,
\newblock ``Efficientnetv2: Smaller models and faster training,''
\newblock in {\em Proceedings of the 38th International Conference on Machine
  Learning}, M~Meila and T~Zhang, Eds. 18--24 Jul 2021, vol. 139 of {\em
  Proceedings of Machine Learning Research}, pp. 10096--10106, PMLR.

\bibitem{rxte}
``Rxte public data,'' \url{ https://heasarc.gsfc.nasa.gov/docs/xte/xte
  public.html}.

\bibitem{adigoke}
O~Adegoke, P~Dhang, B~Mukhopadhyay, M.~C Ramadevi, and D~Bhattacharya,
\newblock ``{Correlating non-linear properties with spectral states of RXTE
  data: possible observational evidences for four different accretion modes
  around compact objects},''
\newblock {\em Monthly Notices of the Royal Astronomical Society}, vol. 476,
  no. 2, pp. 1581--1595, 02 2018.

\end{thebibliography}

\end{document}